\documentclass{article}

\usepackage[utf8]{inputenc}
\usepackage{spconf}

\usepackage[hidelinks, pdfusetitle]{hyperref}

\usepackage{amsmath}
\usepackage{caption}
\usepackage{float}
\usepackage{graphicx}
\usepackage{placeins}
\usepackage{subcaption}
\usepackage{xcolor}
\usepackage{xfrac}
\usepackage{url}

\usepackage[capitalize, nameinlink]{cleveref}

\usepackage{flushend}
\usepackage{fancyhdr}
\fancypagestyle{firstpage}{
  \fancyhf{}
  \fancyhead[C]{\vspace{-2.2cm}Extended version of the paper ``Latent Space Motion Analysis for Collaborative Intelligence,'' to be presented at the IEEE International Conference on Acoustics, Speech and Signal Processing (ICASSP), Toronto, Canada, June 6-11, 2021. \vspace{0.3cm}} 
}

\title{Analysis of Latent-Space Motion for Collaborative Intelligence}
\author{Mateen Ulhaq, Ivan V. Baji\'c}

\name{Mateen Ulhaq and Ivan V. Baji\'c\thanks{This work was supported in part by the Natural Sciences and Engineering Council (NSERC) of Canada.}}
\address{%
  School of Engineering Science,
  Simon Fraser University, Burnaby, BC, Canada%
}

\crefname{equation}{}{}

\DeclareMathOperator{\sign}{sign}

\begin{document}

\maketitle

\begin{abstract}
When the input to a deep neural network (DNN) is a video signal, a sequence of feature tensors is produced at the intermediate layers of the model. If neighboring frames of the input video are related through motion, a natural question is, ``what is the relationship between the corresponding feature tensors?'' By analyzing the effect of common DNN operations on optical flow, we show that the motion present in each channel of a feature tensor is approximately equal to the scaled version of the input motion. The analysis is validated through experiments utilizing common motion models. 
\end{abstract}

\begin{keywords}
Collaborative intelligence, latent space motion, feature domain motion, feature compression 
\end{keywords}

\thispagestyle{firstpage}

\section{Introduction}

\begin{figure}
    \centering
    \includegraphics[width=\columnwidth]{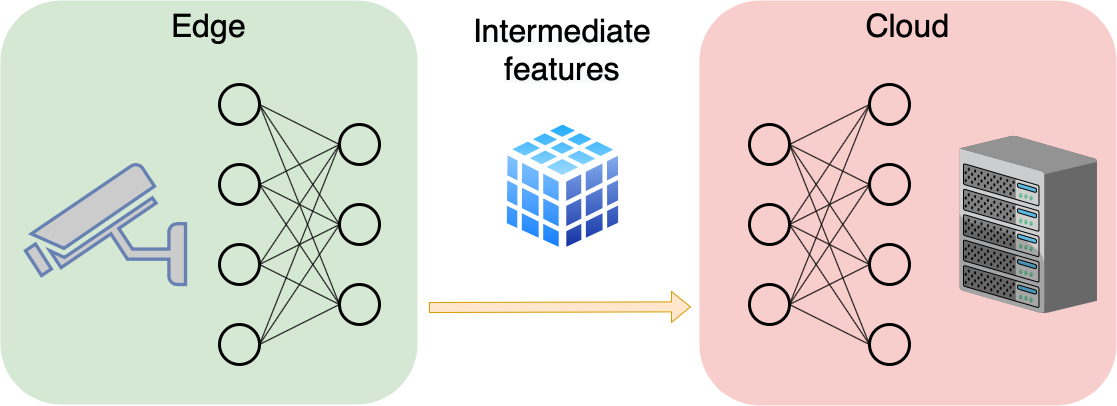}
    \caption{Basic collaborative intelligence system.}
    \label{fig:CI_system}
\end{figure}

Collaborative intelligence (CI)~\cite{Bajic_etal_ICASSP21} has emerged as a promising strategy to bring AI ``to the edge.'' In a typical CI system (\cref{fig:CI_system}), a deep neural network (DNN) is split into two parts: the edge sub-model, deployed on the edge device near the sensor, and the cloud sub-model deployed in the cloud. Intermediate features produced by the edge sub-model are transferred from the edge device to the cloud. It has been shown that such a strategy may provide better energy efficiency~\cite{kang2017neurosurgeon,jointdnn}, lower latency~\cite{kang2017neurosurgeon,jointdnn,ulhaq2019neurips_demo}, and lower bitrates over the communication channel~\cite{dfc_for_collab_object_detection,Choi2018NearLosslessDF}, compared to more traditional cloud-based analytics where the input signal is directly sent to the cloud. These potential benefits will find a number of applications in areas such as intelligent sensing~\cite{Chen19} and video coding for machines~\cite{MPEG_VCM_CFE,Duan2020VideoCF}. In particular, compression of intermediate features has become an important research problem, with a number of recent developments~\cite{Saeed_ICIP19,Hyomin_ICASSP20,Saeed_ICASSP20,Bob_ICME20, Saeed2020pareto_arxiv} for the case when the input to the edge sub-model is a still image.

\begin{figure}
    \centering
    \includegraphics[width=\columnwidth]{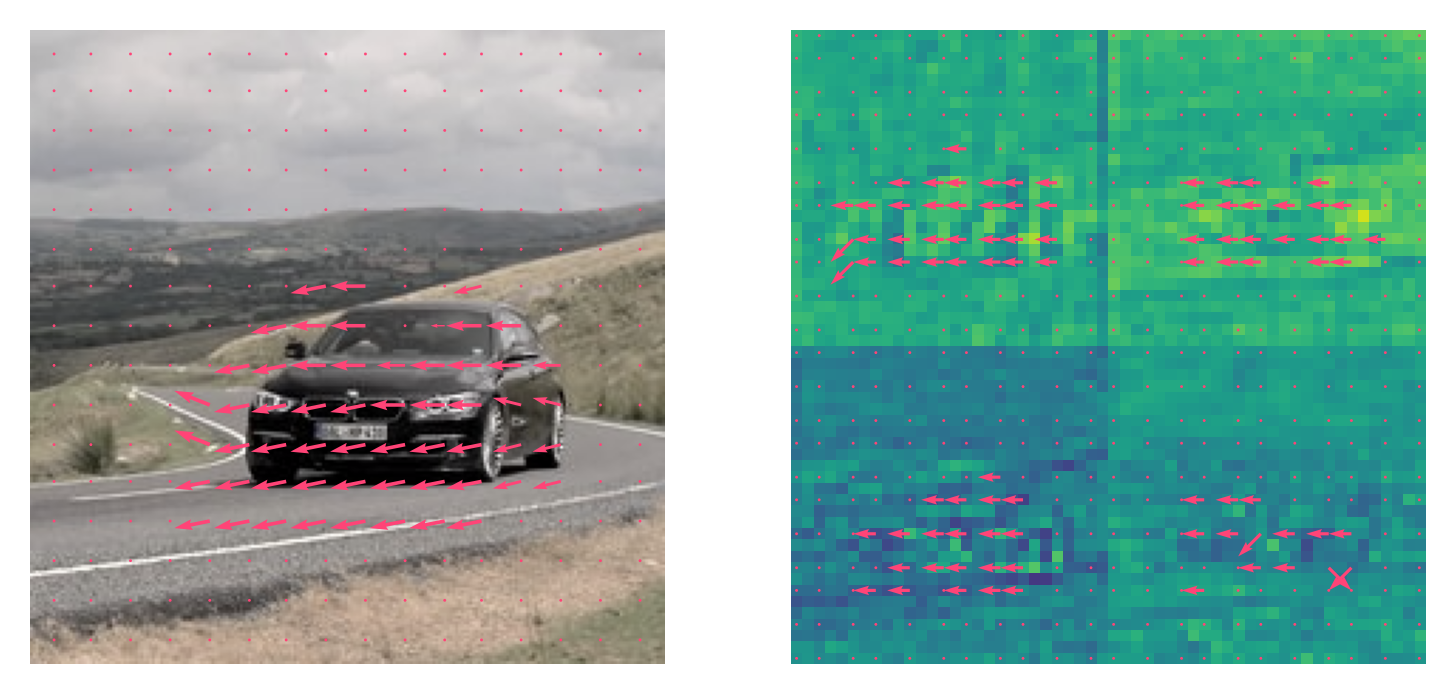}
    \caption{%
        Motion estimates for input frames (left) and
        select channels from the output of ResNet-34's add\_3 layer (right).
    }
    \label{fig:mv_overview}
    \vspace{-1.0\baselineskip}
\end{figure}

When the input to the edge sub-model is video, its output is a sequence of feature tensors produced from successive frames in the input video. This sequence of feature tensors needs to be compressed prior to transmission and then decoded in the cloud for further processing. Since motion plays such an important role in video processing and compression, we are motivated to examine whether any similar relationship exists in the latent space among the feature tensors. Our theoretical and experimental results show that, indeed, motion from the input video is approximately preserved in the channels of the feature tensor. 
An illustration of this is presented in \cref{fig:mv_overview}, where the estimated input-space motion field is shown on the left, and the estimated motion fields in several feature tensor channels are shown on the right.
These findings suggest that methods for motion estimation, compensation, and analysis that have been developed for conventional video processing and compression may provide a solid starting point for equivalent operations in the latent space.   

The paper is organized as follows. In \cref{sec:motion_analysis}, we analyze the actions of typical operations found in deep convolutional neural networks on optical flow in the input signal, and show that these operations tend to preserve the optical flow, at least approximately, with an appropriate scale. In \cref{sec:experiments} we provide empirical support for the theoretical analysis from \cref{sec:motion_analysis}. 
Finally, \cref{sec:conclusions} concludes the paper.

\FloatBarrier

\begin{figure}
    \centering
    \includegraphics[width=0.7\linewidth]{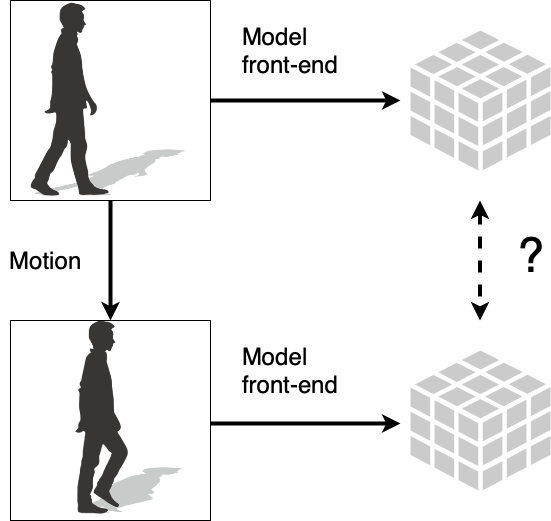}
    \caption{The problem studied in this paper: if input images are related via motion, what is the relationship between the corresponding intermediate feature tensors?
    }
    \label{fig:overview}
\end{figure}









\section{Latent space motion analysis}
\label{sec:motion_analysis}

The basic problem studied in this paper is illustrated in \cref{fig:overview}. Consider two images (video frames) input to the edge sub-model of a CI system. It is common to represent their relationship via a motion model. The question we seek to answer here is, ``what is the relationship between the corresponding feature tensors produced by the edge sub-model?'' To answer this question, we will look at the processing pipeline between the input image and a given channel of a feature tensor. In most deep models for computer vision applications, this processing pipeline consists of a sequence of basic operations: convolutions, pointwise nonlinearities, and pooling.  We will show that each of these operations tends to preserve motion, at least approximately, in a certain sense, and from this we will conclude that (approximate) input motion may be observed in individual channels of a feature tensor.   

\textbf{Motion model.} Optical flow is a frequently used motion model in computer vision and video processing. In a ``2D+t'' model, $I(x,y,t)$ denotes pixel intensity at time $t$, at spatial position $(x,y)$. Under a constant-intensity assumption,  optical flow satisfies the following partial differential equation~\cite{Horn_Schunk_1981}:
\vspace{-5pt}
\begin{equation}
    \frac{\partial I}{\partial x}v_x + \frac{\partial I}{\partial y}v_y + \frac{\partial I}{\partial t} = 0,
    \label{eq:2D+t_opt_flow}
\end{equation}
where $(v_x,v_y)$ represents the motion vector. For notational simplicity, in the analysis below we will use a ``1D+t'' model, which captures all the main ideas but keeps the equations shorter. In a ``1D+t'' model, $I(x,t)$ denotes intensity at position $x$ at time $t$, and the optical flow equation is
\begin{equation}
    \frac{\partial I}{\partial x}v  + \frac{\partial I}{\partial t} = 0,
    \label{eq:1D+t_opt_flow}
\end{equation}
with $v$ representing the motion. 
We will analyze the effect of basic operations --- convolutions, pointwise nonlinearities, and pooling --- on \cref{eq:1D+t_opt_flow}, to gain insight into the relationship between input space motion and latent space motion.

\textbf{Convolution.} Let $f$ be a (spatial) filter kernel, then the optical flow after convolution is a solution to the following equation
\vspace{-5pt}
\begin{equation}
    \frac{\partial}{\partial x}(f*I)v'  + \frac{\partial}{\partial t}(f*I) = 0,
    \label{eq:1D+t_conv}
\end{equation}
where $v'$ is the motion after the convolution. Since the convolution and differentiation are linear operations, we have
\vspace{-3pt}
\begin{equation}
    f*\left(\frac{\partial I}{\partial x}v'  + \frac{\partial I}{\partial t} \right) = 0.
    \label{eq:1D+t_conv_2}
\vspace{-2pt}
\end{equation}
Hence, solution $v$ from~\cref{eq:1D+t_opt_flow} is also a solution of~\cref{eq:1D+t_conv_2}, but~\cref{eq:1D+t_conv_2} could also have other solutions, besides those that satisfy~\cref{eq:1D+t_opt_flow}. 

\textbf{Pointwise nonlinearity.} Nonlinear activations such as \texttt{sigmoid}, \texttt{ReLU}, etc., are commonly applied in a pointwise fashion on the output of convolutions in deep models. Let $\sigma(\cdot)$ denote such a pointwise nonlinearity, then the optical flow after this nonlinearity is a solution to the following equation
\vspace{-3pt}
\begin{equation}
    \frac{\partial}{\partial x}\left[\sigma(I)\right]v'  + \frac{\partial}{\partial t}\left[\sigma(I)\right] = 0,
    \label{eq:1D+t_nonlin}
\end{equation}
where $v'$ is the motion after the pointwise nonlinearity. By using the chain rule of differentiation, the above equation can be rewritten as
\vspace{-7pt}
\begin{equation}
    \sigma'(I)\cdot \left(\frac{\partial I}{\partial x}v'  + \frac{\partial I}{\partial t} \right) = 0.
    \label{eq:1D+t_nonlin_2}
\end{equation}
Hence, again, solution $v$ from~\cref{eq:1D+t_opt_flow} is also a solution of~\cref{eq:1D+t_nonlin_2}. It should be noted that~\cref{eq:1D+t_nonlin_2} may have solutions other than those from~\cref{eq:1D+t_opt_flow}. For example, in the region where inputs to \texttt{ReLU} are negative, the corresponding outputs will be zero, so $\sigma'(I)=0$. Hence, in those regions,~\cref{eq:1D+t_nonlin_2} will be satisfied for arbitrary $v'$. Nonetheless, the solution from~\cref{eq:1D+t_opt_flow} is still one of those arbitrary solutions. 

\textbf{Pooling.} There are various forms of pooling, such as max-pooling, mean-pooling, learnt pooling (via strided convolutions), etc. All these can be decomposed into a sequence of two operations: a spatial operation (local maximum or convolution) followed by scale change (downsampling). Spatial convolution operations can be analyzed as above, and the conclusion is that motion before such an operation is also a solution to the optical flow equation after such an operation. Hence, we will focus here on the local maximum operation and the scale change. 

\textit{Local maximum.} Consider the maximum of function $I(x,t)$ over a local spatial region $[x_0-h, x_0+h]$, at a given time $t$. We can approximate $I(x,t)$ as a locally-linear function, whose slope is the spatial derivative of $I$ at $x_0$, $\frac{\partial}{\partial x}I(x_0,t)$. If the derivative is positive, the maximum is $I(x_0+h,t)$, and if it is negative, it is $I(x_0-h,t)$. In the special case when the derivative is zero, any point in $[x_0-h, x_0+h]$, including the endpoints, is a maximum. From  Taylor series expansion of $I(x,t)$ around $x_0$ up to and including the first-order term,
\vspace{-5pt}
\begin{equation}
    I(x_0 \pm \epsilon,t) \approx I(x_0,t) \pm \frac{\partial}{\partial x}I(x_0,t)\cdot \epsilon,
\end{equation}
for $|\epsilon| \leq h$. With such linear approximation, the local maximum of $I(x,t)$ over $[x_0-h, x_0+h]$ occurs either at $x_0+h$ or at $x_0-h$, depending on the sign of $\frac{\partial}{\partial x}I(x_0,t)$; if the derivative is zero, every point in the interval is a local maximum. Hence, the local maximum of $I(x,t)$ can be approximated as
\vspace{-12pt}
\begin{equation}
\begin{split}
    &\max_{x\in[x_0-h, x_0+h]} I(x,t) \\
    & \qquad \approx I(x_0,t) + \sign\left(\frac{\partial}{\partial x}I(x_0,t)\right) \cdot \frac{\partial}{\partial x}I(x_0,t) \cdot h.
\end{split}
\vspace{-6pt}
\label{eq:local_max}
\end{equation}
Let~\cref{eq:local_max} be the definition of $M(x_0,t)$, the function that takes on local spatial maximum values of $I(x,t)$ over windows of size $2h$. The optical flow after such a local maximum operation is described by
\vspace{-3pt}
\begin{equation}
    \frac{\partial M}{\partial x}v'  + \frac{\partial M}{\partial t} = 0,
    \label{eq:1D+t_local_max}
\end{equation}
where $v'$ represents the motion after local spatial maximum operation. Using~\cref{eq:local_max} in~\cref{eq:1D+t_local_max}, after some manipulation we obtain the following equation
\begin{equation}
\begin{split}
    \frac{\partial I}{\partial x}v'  + \frac{\partial I}{\partial t} + \sign\left(\frac{\partial I}{\partial x}\right) \cdot \frac{\partial}{\partial x} \left( \frac{\partial I}{\partial x}v'  + \frac{\partial I}{\partial t}\right)\cdot h = 0.
\end{split}
\label{eq:1D+t_post_max}
\end{equation}
Note that if $v'$ satisfies the original optical flow equation~\cref{eq:1D+t_conv}, it will also satisfy~\cref{eq:1D+t_post_max}, hence pre-max motion $v$ is also one possible solution to post-max motion $v'$.

\textit{Scale change.} Finally, consider the change of spatial scale by a factor $s$, such that the new signal is $I'(x,t) =I(s\cdot x,t)$. The optical flow equation is now
\begin{equation}
    \frac{\partial I'}{\partial x}v'  + \frac{\partial I'}{\partial t} = 0.
    \label{eq:1D+t_scale_change}
\end{equation}
Since $\frac{\partial I'}{\partial x}=s\cdot \frac{\partial I}{\partial x}$ and $\frac{\partial I'}{\partial t}=\frac{\partial I}{\partial t}$, we conclude that $v'=v/s$, where $v$ is the solution to pre-scaling motion~\cref{eq:1D+t_opt_flow}. Hence, as expected, down-scaling the signal spatially by a factor of $2$ ($s=2$) would reduce the motion by a factor of $2$.

Combining the results of the above analyses, we conclude that convolutions, pointwise nonlinearities, and local maximum operations tend to be motion-preserving operations, in the sense that pre-operation motion is also a solution to post-operation optical flow, at least approximately. The operation with the most obvious impact on motion is scale change. Hence, when looking at latent-space motion at some layer in a deep model, we should expect to find motion similar to the input motion, but scaled down by a factor of $n^k$, where $k$ is the number of pooling operations (over $n\times n$ windows) between the input and the layer of interest. Specifically, if $\mathbf{v}$ is the motion vector at some position in the input frame, then at the corresponding spatial location in all the channels of the feature tensor we can expect to find the vector
\begin{equation}
    \mathbf{v}'\approx \mathbf{v}/n^k.
    \label{eq:mv_scale}
\end{equation}
In \cref{sec:experiments}, we will verify these conclusions experimentally.

\section{Experiments}
\label{sec:experiments}

\begin{figure}[t]
    \centering
    \includegraphics[width=0.95\linewidth]{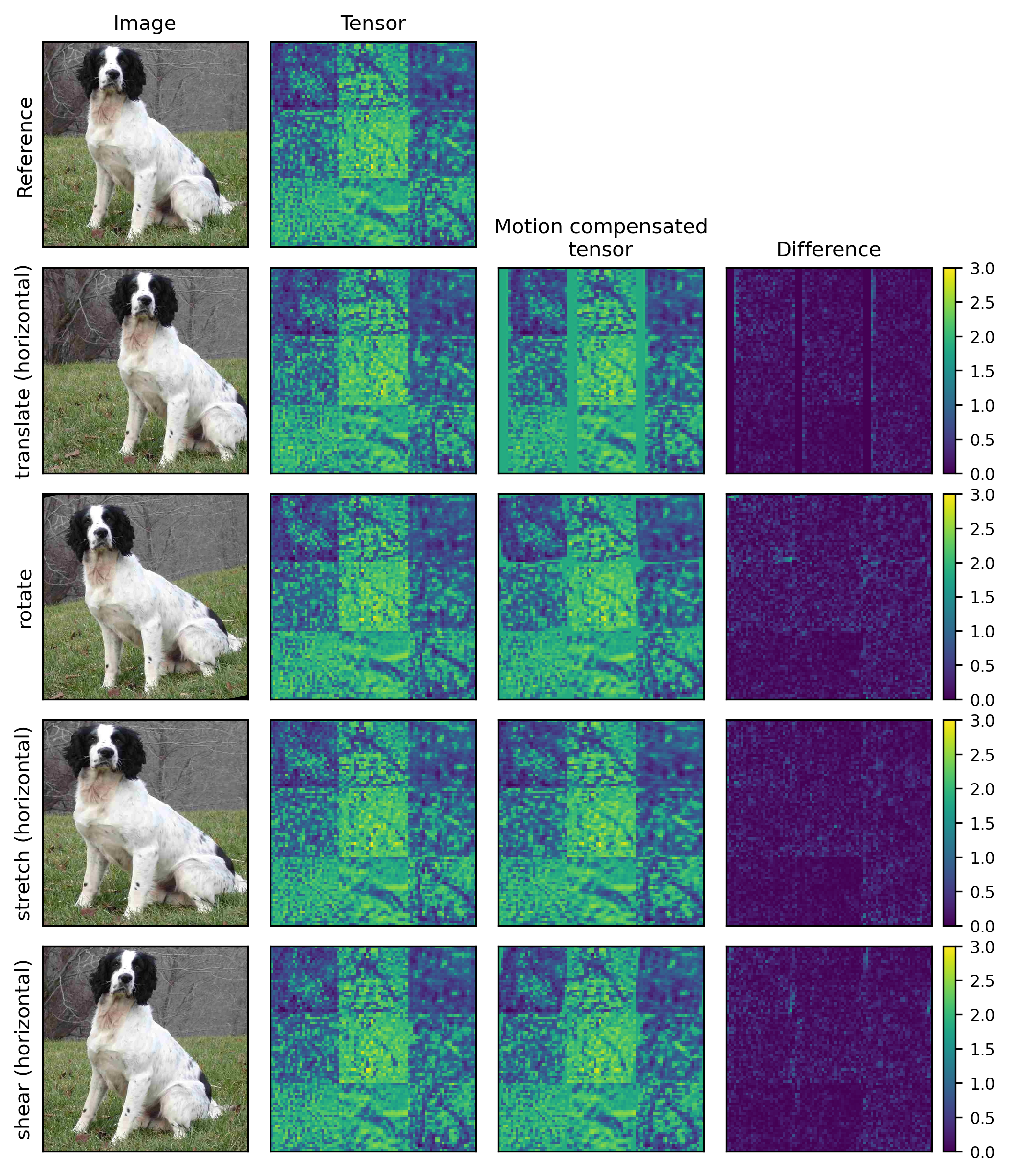}
    \caption{%
        Examples of motion transformations applied to reference image.
        The output tensors of ResNet-34's add\_3 layer are reliably predicted from only the reference tensor and known input-space motion. 
        %
        %
    }
    \label{fig:examples}
    \vspace{-1.0\baselineskip}
\end{figure}

\begin{figure*}[tb]
    \centering
    \includegraphics[width=0.9\linewidth]{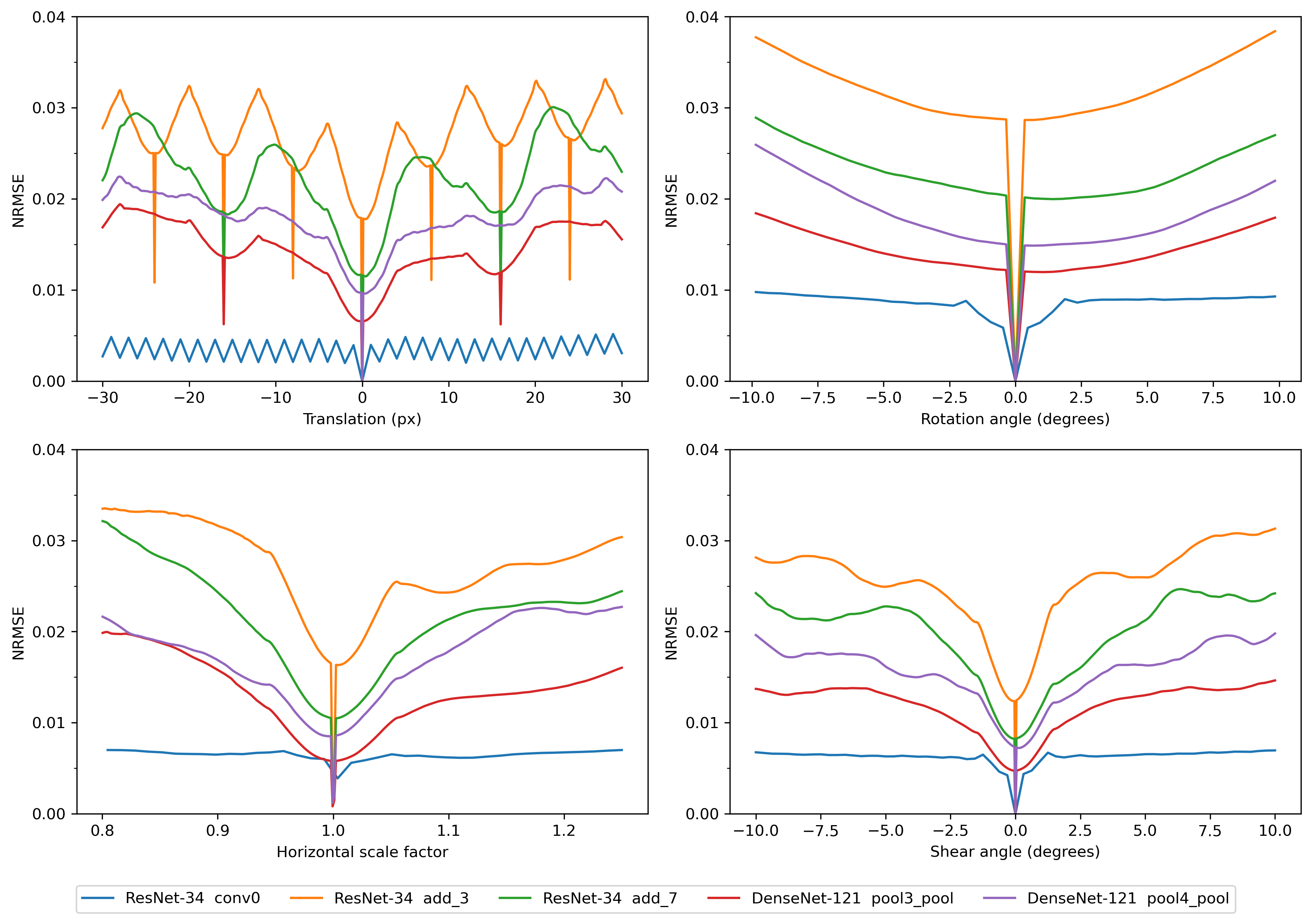}
    \caption{%
        NRMSE across parameter ranges for translation (top-left), rotation (top-right), scaling (bottom-left), and shear (bottom-right).
        For translation, NRMSE local minima occur when the input-space shifts correspond to integer latent-space shifts in~\cref{eq:mv_scale}.%
    }
    \label{fig:transform}
\end{figure*}

\begin{figure}
    \centering
    \includegraphics[width=\linewidth]{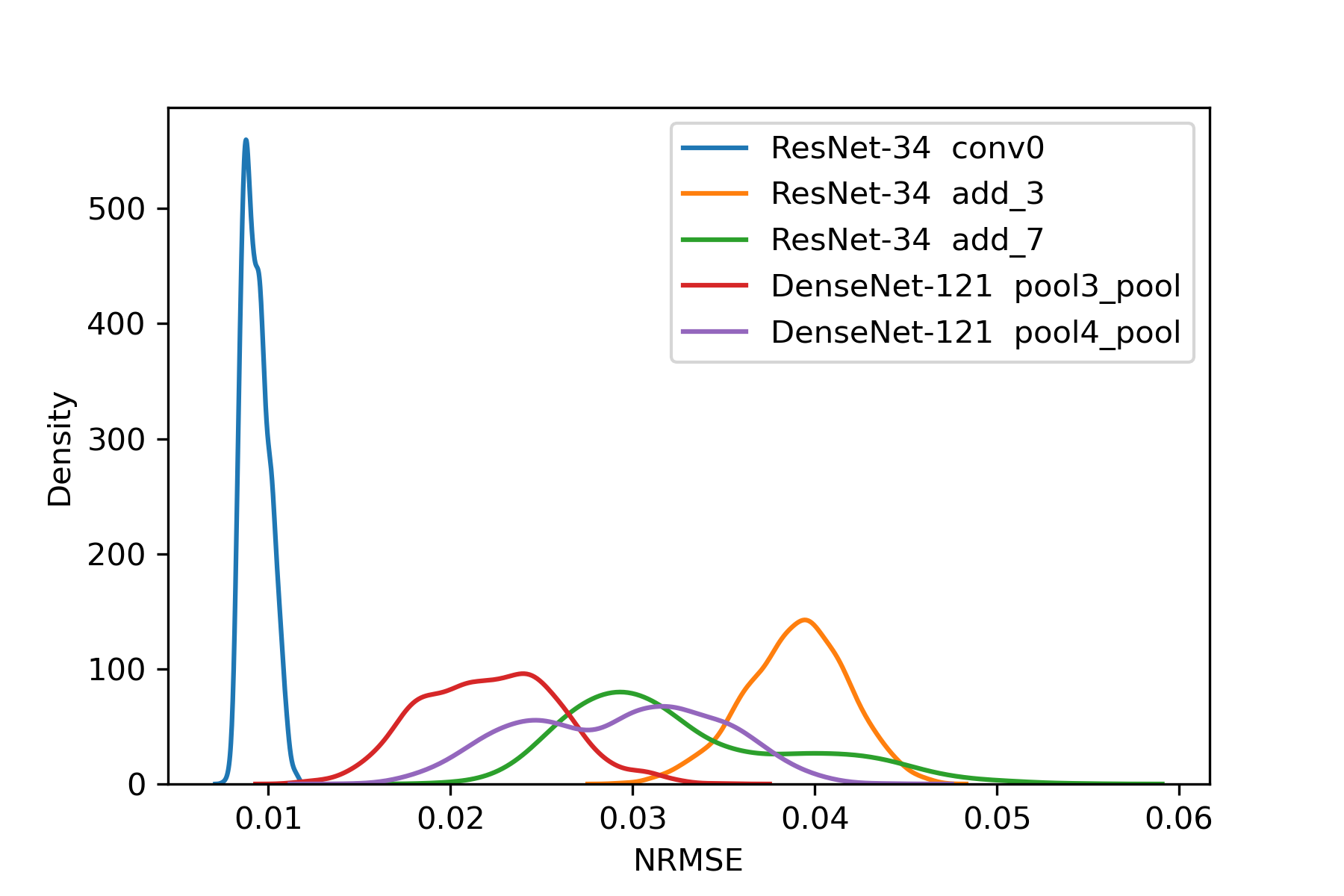}
    \caption{%
        NRMSE histogram computed for an affine motion model~\cite{wang_etal_2002} over the combination of the following seven independent uniformly distributed parameters: x- and y-translation ($\pm 32$ px), x- and y-scaling (0.95--1.05), x- and y-shearing ($\pm 5^{\circ}$), and rotation ($\pm 10^{\circ}$).%
    }
    \label{fig:mse/histogram}
\end{figure}

%
%
%

An illustration of the correspondence between the input-space motion and latent-space motion was shown in \cref{fig:mv_overview}. This example was produced using a pair of frames from a video of a moving car.  The motion vectors were estimated using an exhaustive block-matching search at each pixel, which sought to minimize the sum of squared differences (SSD). In the input frames, whose resolution was $224 \times 224$, the block size of $31 \times 31$ around each pixel and 
the search range of $\pm11$ were used. In the corresponding feature tensor channels, whose resolution was $28 \times 28$, the block size of $3 \times 3$ and a search range of $\pm5$ were used. Although the estimated motion vector fields are somewhat noisy, the similarity between the input-space motion and latent-space motion is evident. 

To examine the relationship between input-space and latent-space motion more closely, we performed several experiments with synthetic input-space motion. In this case, exact input-space motion is known, so relationship~\cref{eq:mv_scale} can be tested more reliably.  \cref{fig:examples} shows examples of various transformations (translation, rotation, stretching, shearing) applied to an input image of a dog. The second column displays several channels from the actual tensor produced by the transformed image, and the third column shows the corresponding channels produced by motion compensating the tensor of the original image via~\cref{eq:mv_scale}. The last column shows the difference between the actual and predicted tensor channels. Note that regions that cannot be predicted, such as regions ``entering the frame,'' were excluded from difference computation. As seen in \cref{fig:examples}, the model~\cref{eq:mv_scale} works reasonably well, and the differences between the actual and predicted tensors are low.



For quantitative evaluation, experiments were conducted on several layers of ResNet-34~\cite{ResNet} and DenseNet-121~\cite{DenseNet}. Normalized Root Mean Square Error (NRMSE)~\cite{wiki:NRMSE} was used for this purpose:
\vspace{-7pt}
\begin{equation}
    \text{NRMSE} =  \frac{1}{R} \sqrt{\frac{1}{N} \sum_{i=1}^N (p_i - a_i)^2},
    \label{eq:NRMSE}
\end{equation}
where $a_i$ is the actual tensor value produced 
from the transformed input, $p_i$
is the tensor value predicted using our motion model~\cref{eq:mv_scale}, $N$ is the number of elements in the feature tensor, and $R$ is the dynamic range. Again, regions that cannot be predicted were excluded from NRMSE computation. \cref{fig:transform} shows NRMSE computed across a range of parameters for several transformations, at various layers of the two DNNs. 





As seen in \cref{fig:transform}, NRMSE goes up to about 0.04 for reasonable ranges of transformation parameters. How good is this? To answer this question, we set out to find the typical values of NRMSE found in conventional motion-compensated frame prediction.
In a recent study~\cite{Choi_TCSVT_2020}, the quality of frames predicted by conventional motion estimation and motion compensation (MEMC) in High Efficiency Video Coding (HEVC)~\cite{H.265} was compared against a DNN developed for frame prediction. From Table~III in~\cite{Choi_TCSVT_2020}, the luminance Peak Signal to Noise Ratio (PSNR) of frames predicted uni-directionally by the DNN and conventional HEVC MEMC was in the range 27--41 dB over several HEVC test sequences. NRMSE can be computed from PSNR as
\vspace{-2pt}
\begin{equation}
    \text{NRMSE} = \frac{1}{256}\sqrt{\frac{255^2}{10^{\text{PSNR}/10}}},
\vspace{-2pt}
\end{equation}
so the PSNR range of 27--41 dB corresponds to the NRMSE range of 0.009--0.044. These levels of NRMSE are indicative of how much motion models used in video coding deviate from the true motion. As seen in \cref{fig:transform}, the model~\cref{eq:mv_scale} produces NRMSE in the same range, so the accuracy of~\cref{eq:mv_scale} is comparable to the accuracy of common motion models used in video coding. Another illustration of this is presented in \cref{fig:mse/histogram}, which shows the histogram of NRMSE computed across a range of affine transformation parameters.  Hence,~\cref{eq:mv_scale} represents a good starting point for development of latent-space motion compensation.

\section{Conclusions}
\label{sec:conclusions}
Using the concept of optical flow, in this paper we analyzed motion in the latent space of a deep model induced by the motion in the input space, and showed that motion tends to be approximately preserved in the channels of intermediate feature tensors. These findings suggest that motion estimation, compensation, and analysis methods developed for conventional video signals should be able to provide a good starting point for latent-space motion processing, such as motion-compensated prediction and compression, tracking, action recognition, and other applications. 

\bibliographystyle{IEEEbib}
\bibliography{references}

\begin{thebibliography}{10}

\bibitem{Bajic_etal_ICASSP21}
I.~V. Baji\'{c}, W.~Lin, and Y.~Tian,
\newblock ``Collaborative intelligence: Challenges and opportunities,''
\newblock in {\em Proc. IEEE ICASSP}, 2021,
\newblock to appear.

\bibitem{kang2017neurosurgeon}
Y.~Kang, J.~Hauswald, C.~Gao, A.~Rovinski, T.~Mudge, J.~Mars, and L.~Tang,
\newblock ``Neurosurgeon: Collaborative intelligence between the cloud and
  mobile edge,''
\newblock in {\em Proc. 22nd ACM Int. Conf. Arch. Support Programming Languages
  and Operating Syst.}, 2017, pp. 615--629.

\bibitem{jointdnn}
A.~E. Eshratifar, M.~S. Abrishami, and M.~Pedram,
\newblock ``{JointDNN}: An efficient training and inference engine for
  intelligent mobile cloud computing services,''
\newblock {\em IEEE Trans. Mobile Computing}, 2019,
\newblock Early Access.

\bibitem{ulhaq2019neurips_demo}
M.~Ulhaq and I.~V. Baji\'{c},
\newblock ``Shared mobile-cloud inference for collaborative intelligence,''
\newblock {\em arXiv:2002.00157}, 2019,
\newblock NeurIPS'19 demonstration.

\bibitem{dfc_for_collab_object_detection}
H.~Choi and I.~V. Baji\'{c},
\newblock ``Deep feature compression for collaborative object detection,''
\newblock in {\em Proc. IEEE ICIP}, Oct. 2018, pp. 3743--3747.

\bibitem{Choi2018NearLosslessDF}
H.~Choi and I.~V. Baji\'{c},
\newblock ``Near-lossless deep feature compression for collaborative
  intelligence,''
\newblock in {\em Proc. IEEE MMSP}, Aug. 2018, pp. 1--6.

\bibitem{Chen19}
Z.~Chen, K.~Fan, S.~Wang, L.~Duan, W.~Lin, and A.~C. Kot,
\newblock ``Toward intelligent sensing: Intermediate deep feature
  compression,''
\newblock {\em IEEE Trans. Image Processing}, vol. 29, pp. 2230--2243, 2019.

\bibitem{MPEG_VCM_CFE}
ISO/IEC,
\newblock ``Draft call for evidence for video coding for machines,''
\newblock ISO/IEC JTC 1/SC 29/WG 11 W19508, Jul. 2020.

\bibitem{Duan2020VideoCF}
L.~Duan, J.~Liu, W.~Yang, T.~Huang, and W.~Gao,
\newblock ``Video coding for machines: A paradigm of collaborative compression
  and intelligent analytics,''
\newblock {\em IEEE Transactions on Image Processing}, vol. 29, pp. 8680--8695,
  2020.

\bibitem{Saeed_ICIP19}
S.~R. Alvar and I.~V. Baji\'{c},
\newblock ``Multi-task learning with compressible features for collaborative
  intelligence,''
\newblock in {\em Proc. IEEE ICIP}, Sep. 2019, pp. 1705--1709.

\bibitem{Hyomin_ICASSP20}
H.~Choi, R.~A. Cohen, and I.~V. Baji\'{c},
\newblock ``Back-and-forth prediction for deep tensor compression,''
\newblock in {\em Proc. IEEE ICASSP}, 2020, pp. 4467--4471.

\bibitem{Saeed_ICASSP20}
S.~R. Alvar and I.~V. Baji\'{c},
\newblock ``Bit allocation for multi-task collaborative intelligence,''
\newblock in {\em Proc. IEEE ICASSP}, May 2020, pp. 4342--4346.

\bibitem{Bob_ICME20}
R.~A. Cohen, H.~Choi, and I.~V. Baji\'{c},
\newblock ``Lightweight compression of neural network feature tensors for
  collaborative intelligence,''
\newblock in {\em Proc. IEEE ICME}, Jul. 2020, pp. 1--6.

\bibitem{Saeed2020pareto_arxiv}
S.~R. Alvar and I.~V. Baji\'{c},
\newblock ``Pareto-optimal bit allocation for collaborative intelligence,''
\newblock {\em arXiv:2009.12430}, Sep. 2020.

\bibitem{Horn_Schunk_1981}
B.~K.~P. Horn and B.~G. Schunck,
\newblock ``Determining optical flow,''
\newblock {\em Artificial Intelligence}, vol. 17, no. 1, pp. 185 -- 203, 1981.

\bibitem{wang_etal_2002}
Y.~Wang, J.~Ostermann, and Y.-Q. Zhang,
\newblock {\em Video Processing and Communications},
\newblock Prentice-Hall, 2002.

\bibitem{ResNet}
K.~{He}, X.~{Zhang}, S.~{Ren}, and J.~{Sun},
\newblock ``Deep residual learning for image recognition,''
\newblock in {\em Proc. IEEE CVPR}, 2016, pp. 770--778.

\bibitem{DenseNet}
G.~{Huang}, Z.~{Liu}, L.~{Van Der Maaten}, and K.~Q. {Weinberger},
\newblock ``Densely connected convolutional networks,''
\newblock in {\em Proc. IEEE CVPR}, 2017, pp. 2261--2269.

\bibitem{wiki:NRMSE}
{Wikipedia contributors},
\newblock ``Root-mean-square deviation --- {Wikipedia}{,} the free
  encyclopedia,'' 2020,
\newblock [Online] Available:
  https://en.wikipedia.org/wiki/Root-mean-square\_deviation.

\bibitem{Choi_TCSVT_2020}
H.~Choi and I.~V. Baji\'{c},
\newblock ``Deep frame prediction for video coding,''
\newblock {\em IEEE Trans. Circuits Syst. Video Technol.}, vol. 30, no. 7, pp.
  1843--1855, Jul. 2020.

\bibitem{H.265}
ITU,
\newblock ``High efficiency video coding,'' Recommendation ITU-T H.265, Nov.
  2019.

\end{thebibliography}

\end{document}